\documentclass[letterpaper, 10 pt, conference]{IEEEtran}
\usepackage{comment}
\usepackage{siunitx}
\usepackage{relsize}
\usepackage{ifthen}
\usepackage[colorinlistoftodos]{todonotes}

\usepackage{graphics} %
\usepackage{rotating}
\usepackage{color}
\usepackage{enumerate}
\usepackage[T1]{fontenc}
\usepackage{psfrag}
\usepackage{epsfig} %
\usepackage{booktabs}
\usepackage{graphicx,url}
\usepackage{multirow}
\usepackage{array}
\usepackage{latexsym}
\usepackage{amsfonts}
\usepackage{amsmath}
\usepackage{amssymb}
\usepackage{xstring}
\usepackage{multirow}
\usepackage{xcolor}
\usepackage{prettyref}
\usepackage{flexisym}
\usepackage{bigdelim}
\usepackage{breqn} %
\usepackage{listings}
\let\labelindent\relax
\usepackage{enumitem}
\usepackage{xspace}
\usepackage{bm}
\graphicspath{{./figures/}}
\usepackage{tikz}
\usetikzlibrary{matrix,calc}

\usepackage{mdwlist}

\let\itemize\undefined
\makecompactlist{itemize}{stditemize}

\usepackage{etex}
\usepackage{algpseudocode}
\usepackage{algorithm}
\usepackage{subcaption}
\usepackage{amsmath}

\DeclareCaptionLabelSeparator{periodspace}{.\quad}
\IEEEoverridecommandlockouts  

\usepackage[letterpaper, left=0.75in, right=0.75in, bottom=0.75in, top=0.75in]{geometry}
\usepackage{multicol}
\usepackage[bookmarks=true]{hyperref}
\usepackage{amsthm}
\newtheoremstyle{mystyle}%
  {}%
  {}%
  {\itshape}%
  {}%
  {\bfseries}%
  {.}%
  { }%
  {\thmname{#1}\thmnumber{ #2}\thmnote{ (#3)}}%
\theoremstyle{mystyle}
\newrefformat{prob}{Problem\,\ref{#1}}
\newrefformat{def}{Definition\,\ref{#1}}
\newrefformat{sec}{Section\,\ref{#1}}
\newrefformat{sub}{Section\,\ref{#1}}
\newrefformat{prop}{Proposition\,\ref{#1}}
\newrefformat{app}{Appendix\,\ref{#1}}
\newrefformat{alg}{Algorithm\,\ref{#1}}
\newrefformat{cor}{Corollary\,\ref{#1}}
\newrefformat{thm}{Theorem\,\ref{#1}}
\newrefformat{lem}{Lemma\,\ref{#1}}
\newrefformat{fig}{Fig.\,\ref{#1}}
\newrefformat{tab}{Table\,\ref{#1}}

\newcommand{\bdmath}{\begin{dmath}}
\newcommand{\edmath}{\end{dmath}}
\newcommand{\beq}{\begin{equation}}
\newcommand{\eeq}{\end{equation}}
\newcommand{\bdm}{\begin{displaymath}}
\newcommand{\edm}{\end{displaymath}}
\newcommand{\bea}{\begin{eqnarray}}
\newcommand{\eea}{\end{eqnarray}}
\newcommand{\beal}{\beq \begin{array}{ll}}
\newcommand{\eeal}{\end{array} \eeq}
\newcommand{\beas}{\begin{eqnarray*}}
\newcommand{\eeas}{\end{eqnarray*}}
\newcommand{\ba}{\begin{array}}
\newcommand{\ea}{\end{array}}
\newcommand{\bit}{\begin{itemize}}
\newcommand{\eit}{\end{itemize}}
\newcommand{\ben}{\begin{enumerate}}
\newcommand{\een}{\end{enumerate}}

\newcommand{\setal}{~\emph{et~al.}\xspace}
\newcommand{\eg}{\emph{e.g.,}\xspace}

\newcommand{\myParagraph}[1]{{\bf #1.}\xspace}
\newcommand{\M}[1]{{\bm #1}} %
\renewcommand{\boldsymbol}[1]{{\bm #1}}

\newcommand{\hide}[1]{}

\newcommand{\hiddenText}{{\color{gray} hidden text.}}
\newcommand{\hideWithText}[1]{\hiddenText}

\newcommand{\ME}{\M{E}}

\newcommand{\MR}{\M{R}}

\newcommand{\va}{\boldsymbol{a}} 
 
\newcommand{\vb}{\boldsymbol{b}}

\newcommand{\vo}{\boldsymbol{o}}

\newcommand{\vr}{\boldsymbol{r}}

\newcommand{\vv}{\boldsymbol{v}}
\newcommand{\vt}{\boldsymbol{t}}
\newcommand{\vxx}{\boldsymbol{x}}

\newcommand{\vgamma}{\boldsymbol{\gamma}}

\newcommand{\blue}[1]{{\color{blue}#1}}

\newcommand{\linkToPdf}[1]{\href{#1}{\blue{(pdf)}}}
\newcommand{\linkToPpt}[1]{\href{#1}{\blue{(ppt)}}}
\newcommand{\linkToCode}[1]{\href{#1}{\blue{(code)}}}
\newcommand{\linkToWeb}[1]{\href{#1}{\blue{(web)}}}
\newcommand{\linkToVideo}[1]{\href{#1}{\blue{(video)}}}
\newcommand{\linkToMedia}[1]{\href{#1}{\blue{(media)}}}
\newcommand{\award}[1]{\xspace} %

\usepackage{float}
\usepackage{hyperref}
\usepackage{graphicx}
\usepackage{epstopdf}
\usepackage{epsfig}
\usepackage{amsmath}
\usepackage[capitalise]{cleveref}

\newcommand{\nameP}{VERF-PnP\xspace}
\newcommand{\nameL}{VERF-Light\xspace}

\newcommand{\name}{VERF\xspace}
\newcommand{\colmap}{COLMAP\xspace}
\newcommand{\nerf}{NeRF\xspace}
\newcommand{\probP}{\text{I\kern-0.15em P}}

\newcommand{\plotheight}{6.5cm}

\usepackage{xr}

\title{\vspace{0.25in}\LARGE \bf{\name: Runtime Monitoring of Pose Estimation \\ with Neural Radiance Fields}}

\author{Dominic Maggio,
\thanks{D.\,Maggio is with the Laboratory for 
Information \& Decision Systems, Massachusetts Institute of Technology, Cambridge, MA, USA, 
and is a Draper Scholar with the Perception and Embedded ML Group,
Draper, Cambridge, MA, USA, 
\sf{drmaggio@mit.edu}
}Courtney Mario,
\thanks{C.\,Mario is with the Draper Perception and Embedded ML Group,
Draper, Cambridge, MA, USA,  
\sf{cmario@draper.com}
}Luca Carlone
\thanks{L.\,Carlone is with the Laboratory for 
Information \& Decision Systems, Massachusetts Institute of Technology, Cambridge, MA, USA, 
\sf{lcarlone@mit.edu}
}
\thanks{This work was partially funded by the NASA Flight Opportunities under grant Nos 80NSSC21K0348 and 80NSSC20K0104.}
}

\begin{document}

\maketitle

\begin{abstract}

We present \name, %
a collection of two methods (\nameP and \nameL) 
for providing runtime assurance on the correctness 
of a camera pose estimate of a monocular camera without relying on direct depth measurements. 
We leverage the ability of \nerf (Neural Radiance Fields) to render novel RGB perspectives of a scene. 
We only require as input the camera image whose pose is being estimated, 
an estimate of the camera pose we want to monitor, and 
a \nerf model containing the scene pictured by the camera. %
We can then predict if the pose estimate is within a desired distance from the ground truth and justify 
our prediction with a level of confidence. 
\nameL does this by rendering a viewpoint with NeRF at the estimated pose and estimating its relative offset to the sensor image up to scale. 
Since scene scale is unknown, the approach renders another auxiliary image and reasons over the consistency of the optical flows across the three images.
\nameP takes a different approach by rendering a stereo pair of images with \nerf and utilizing the Perspective-n-Point (PnP) algorithm. 
We evaluate both methods on the LLFF dataset, on data from a Unitree A1 quadruped robot, 
and on data collected from Blue Origin’s sub-orbital New Shepard rocket %
to 
demonstrate the effectiveness of the proposed pose monitoring method across a range of scene scales. We also show monitoring can be 
completed in under half a second on a 3090 GPU.

\end{abstract} %

\section{Introduction}
\label{sec:intro}

Estimating the pose of a camera from a monocular image is a fundamental problem in computer vision. However, limited work has been done 
to independently monitor the accuracy of the estimated pose and detect incorrect estimates without having direct access to depth information of the scene. 
This need is motivated by the growing use 
of monocular camera localization in high-stakes scenarios such as self driving~\cite{Wang19pami-apolloscape}, spacecraft entry decent and landing 
\cite{Johnson17-lvs}, \cite{Lorenz17ac-osirisrex}, \cite{Maggio23scitech-IBAL}, 
and robotics tasks~\cite{Manuelli19-kpam}. 
For instance, the detection of repeatedly incorrect estimates can be used to decide when to alert the user or trigger mitigation measures (\eg performing a safety landing for a drone, or disengaging the autopilot of a self-driving car).

\begin{figure}[hbt]
    \includegraphics[width=0.47\textwidth]{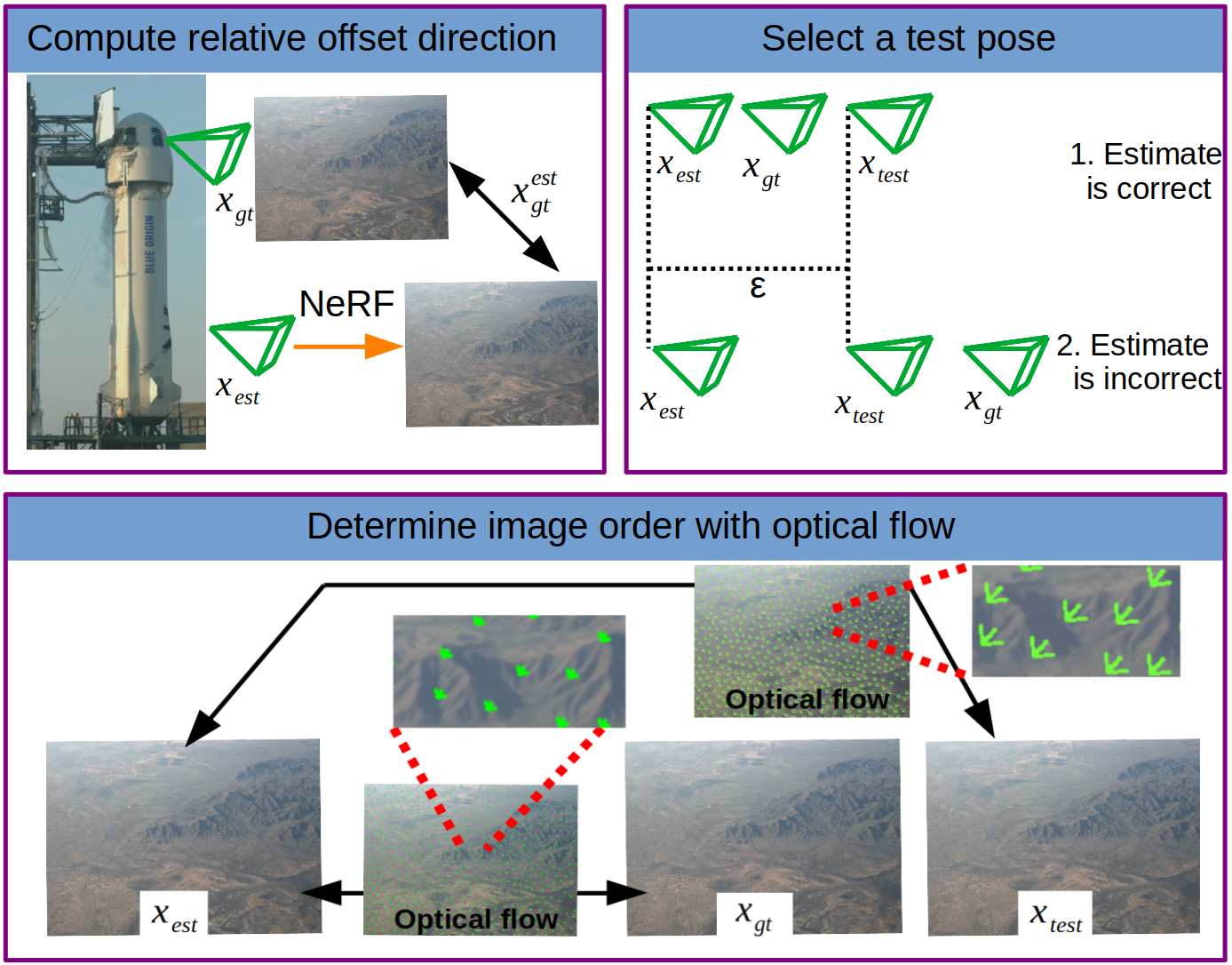}
    \caption{
        Three main phases of \nameL. First, the relative error of a pose estimate up to scale is found by comparing a sensor image (collected at the ground truth pose, $\vxx_{gt}$,) to a NeRF image 
        rendered at the pose estimate, $\vxx_{est}$. Next, a test pose, $\vxx_{test}$, is selected at an $\epsilon$ distance from the estimated pose such that all three poses are co-linear. 
        Determining if the pose estimate is correct is lastly done by estimating the order of the three poses by comparing optical flow between the 
        three corresponding images.
    \label{fig:diagram} \vspace{-6mm}}
\end{figure}

Recent works such as  
\cite{Yen20iros-inerf}, \cite{Adamkiewicz21arxiv-nerfNavigation}, \cite{Maggio23icra-LocNerf}, \cite{Lin22arxiv-NerfInversion}, \cite{Zhu22arxiv-LatitudeNeRF}, \cite{Avraham22cvpr-nerfels} 
explore the use of NeRF~\cite{Mildenhall20arxiv-nerf} 
(Neural Radiance Fields) for camera pose estimation. NeRFs are fully connected networks trained on sequences of RGB images to 
learn an implicit representation of a scene, from which the network can be used to generate RGB images at novel viewpoints. 
As an example, Loc-NeRF~\cite{Maggio23icra-LocNerf} uses NeRF as a map of an environment and utilizes a particle filter backbone to output a pose estimate of a provided sensor image. 
However, there is no clear and reliable measure to determine if the outputted pose is correct ---where we define correct as being within a distance $\epsilon$ of the true pose--- and existing approaches can fail without notice.

To overcome this limitation, we propose \name, a collection of two approaches coined \nameP and \nameL. 
\name uses the sensor image already present in the pose optimization 
phase to provide assurance that the pose estimate is correct. We additionally require a NeRF model of the scene, but NeRF does not need to be used to produce the pose estimate being monitored, which allows 
\name to be used for pose monitoring regardless of the pose estimation method.
\nameP renders a stereo pair of images with NeRF, one of which is at the estimated pose and the other at a given baseline, 
and uses the Perspective-n-Point (PnP) solver 
with RANSAC~\cite{Fischler81} to estimate the relative offset to the sensor image. 
\nameL uses a different methodology
which can be stated concisely as follows. We first render an image with NeRF at the estimated pose, $\vxx_{est}$, and use it to determine the relative translation up to scale 
between the estimated pose and the ground truth pose, $\vxx_{gt}$. To overcome scale ambiguity we render a test image at a pose $\vxx_{test}$ which is at a distance $\epsilon$ from the estimate pose 
in the direction of the sensor image. If the camera origin of these three images are co-linear with no rotation, then we show that we can compare optical flow fields between 
the three images to determine the order of the camera centers and hence the correctness of the pose estimate (\cref{fig:diagram}). 
To enable assurance in 
the presence of noise, we incorporate an estimate of optical flow error and add outlier rejection using geometric constraints 
to compute a measure of confidence instead of a binary decision.
We remark that as the rotation error can be directly observed between the sensor image and the image rendered at the estimated pose, 
we only focus our attention on determining the quality of the position estimate. 
We provide results on the publicly available LLFF dataset~\cite{Mildenhall19tog-llff}, on real data collected by an A1 quadruped,
and on data collected onboard Blue Origin's sub-orbital New Shepard rocket at heights up to 8 km above the ground and at speeds over 800 km/hr. 
The results showcase the potential of \name to perform in challenging real-world conditions.
Our runtime monitoring approach runs in less than half a second on a 3090 GPU. 

The rest of the paper is organized as follows. \cref{sec:related_works} discusses related work. 
\cref{sec:notation} provides notation and preliminary concepts.
Our two approaches are presented in \cref{sec:contribution_pnp,sec:contribution_light}. 
\cref{sec:experiments} evaluates the methods on three types of experiments: LLFF, A1 robot, and sub-orbital rocket.
Finally, \cref{sec:conclusion} concludes the paper. Extra results and studies are included in the appendix (\cref{sec:appendix}). %

\section{Related Work}
\label{sec:related_works}

\myParagraph{Neural Radiance Fields}
NeRF was introduced by Mildenhall\setal~\cite{Mildenhall20arxiv-nerf} and represents a 3D scene with a neural implicit encoding that 
can be used to render novel viewpoints of the scene. Several extensions are beginning to leverage NeRF for robotic tasks such as 
localization. Yen\setal~\cite{Yen20iros-inerf} develop iNeRF which inverted the NeRF paradigm by solving for a pose given an image. 
Adamkiewicz\setal~\cite{Adamkiewicz21arxiv-nerfNavigation} develop NeRF-Navigation which uses NeRF for a full autonomy pipeline of localization, 
planning, and control.  Zhu\setal~\cite{Zhu22arxiv-LatitudeNeRF}  propose LATITUDE to perform pose estimation with large-scale scenes. 
Maggio\setal~\cite{Maggio23icra-LocNerf} develop Loc-NeRF which uses a particle filter backbone and performs localization while using NeRF as a map. 
Lin\setal~\cite{Lin22arxiv-NerfInversion} use parallelized Monte Carlo Sampling to estimate camera 
poses. Rosinol\setal~\cite{Rosinol22arxiv-nerfSLAM} develop NeRF-SLAM which builds a 
NeRF as images and poses become available. Sucar\setal~\cite{Sucar21iccv-iMAP} proposes iMAP and 
Zhu\setal~\cite{Zhu22cvpr-niceslam} develop NICE-SLAM which use depth from a stereo camera along with RGB to create a neural 
implicit map of room-size scenes. Alignment accuracy of NeRF is studied and improved by Jiang el al.~\cite{jiang22arxiv-alignerf}.
Moreau\setal~\cite{Moreau23arxiv-crossfire} develop CROSSFIRE which uses PnP for localization with \nerf by training self-supervised feature descriptors and rendering 
depth directly from a neural renderer. Li\setal~\cite{Li22arxiv-nerfpose} develop NeRF-Pose which uses PnP with \nerf for object pose estimation by training a pose 
regression network to predict 2D-3D correspondences. 

\myParagraph{Visual Localization}
Since \name can monitor the accuracy of a pose estimate independent of the estimation method, we also include a brief mention 
of visual localization methods outside the scope of \nerf. Classical methods for robotic localization typically use 
either matching of sparse keypoints~\cite{Klein07ismar-PTAM, Qin19arxiv-VINS-Fusion-odometry,Cadena16tro-SLAMsurvey} or 
a dense representation~\cite{Newcombe2011iccv-dtam}. 
Visual terrain relative navigation is the problem of estimating the pose 
of a camera given a terrain map (oftentimes built with satellite or aerial imagery and elevation data) 
\cite{Mourikis09tro-EdlSoundingRocket, Johnson17-lvs, Norman22aas_osirisrex, Maggio23scitech-IBAL}. Absolute Pose Regressors~\cite{Sattler19CVPR-CNNLimitationPose} 
use Convolutional Neural Networks to predict poses by learning the end-to-end localization pipeline. 
We refer to 
Piasco\setal~\cite{Piasco18pr-VBLsurvey} for a more in-depth review of visual localization. 

\myParagraph{Certifiable Perception and Runtime Monitoring}
Carlone and Dellaert~\cite{Carlone15icra-verification} and
Rosen\setal~\cite{Rosen18ijrr-sesync} develop optimality certification techniques for 
pose synchronization problems. %
Yang\setal~\cite{Yang20arxiv-teaser, Yang22pami-certifiablePerception} develop certifiable algorithms for outlier robust estimation. %
Garcia-Salguero\setal \cite{Garcia21arXiv-FastCertTwoView, Garcia21IVC-certifiablerelativepose} certify the optimality of a relative pose estimate.
Zhao\setal~\cite{Zhao20cvpr-certifiablyEssential} present a certifiably optimal approach to estimate the generalized essential matrix.
Here, we instead focus on monitoring the correctness of the pose estimate, rather than optimality of the estimation backend.
Yang\setal~\cite{Yang20arxiv-teaser} and Carlone~\cite{Carlone22arxiv-estimationContracts} develop estimation contracts which certify 
the correctness of a geometric perception problem given conditions are met on the inputs. Talak\setal~\cite{Talak23arxiv-c3po} extend 
certification of correctness for learning-based object pose estimation. 
Yang and Pavone~\cite{Yang23arxiv-ransag} provide statistical bounds on object pose estimation given a heatmap predictions of object keypoints. 
Other works provide confidence metrics to 
monitor the correctness of perception algorithms without providing a certificate of correctness. 
Hu and Mordohai~\cite{Hu12pami-stereoConfidenceSurvey} provide a survey on confidence metrics for stereo matching. 
Rahman\setal~\cite{Rahman21access-monitoringSurvey} provide a survey on monitoring the correctness of learning-based methods for 
robotic perception. 
Antonante\setal~\cite{Antonante22arxiv-perceptionMonitoring} use a diagnostic 
graph to formalize detecting and identifying faults in a perception system.

\section{Notation and Preliminaries}
\label{sec:notation}

\myParagraph{Notation}
We use lowercase symbols (e.g., $\epsilon$) to represent scalars, bold lowercase letters (e.g., $\vxx$) for vectors, and 
bold uppercase letters (e.g., $\ME$) for matrices. Sets are represented 
with capital calligraphic fonts (e.g., $\mathcal{R}$). Unit vectors and homogeneous vectors are denoted with a bar 
and tilde (e.g., $\bar{\vxx}$ and $\tilde{\vxx}$) 
respectively. Estimated quantities are shown with a caret (e.g., $\hat{\vxx}$, $\hat{\ME}$). We express the 2-norm of a vector as $\|\cdot\|$.

Let $\vr_i=(x,y)$ be a coordinate in an image $I_i$. The sensor image will be referred to as $I_{gt}$ as it is taken by a camera at the true pose. 
The estimated and test images will be 
referenced as $I_{est}$ and $I_{test}$. Let $\vv(\vr_i)_{I_i,I_j}$ be the optical flow vector at point $\vr$ in some image $i$ to the 
corresponding point in some image $j$ such that 
$\vr_i + \vv(\vr_i)_{I_i,I_j} = \vr_j$. 
$[\va]_\times$ is the skew-symmetric matrix such that $\va \times \vb = [\va]_\times \vb$.

\myParagraph{The Essential Matrix} 
Assuming points have been calibrated using the camera intrinsics, the 
essential matrix $\ME_{i,j}$ relates corresponding homogeneous coordinates $\tilde{\vr}_i$, $\tilde{\vr}_j$ in two images with the following constraint:
\begin{equation}
    \label{eq:essential_matrix_points}
    (\tilde{\vr}_{j})^T \ME_{i,j} \tilde{\vr}_{i} = 0.
\end{equation}

The matrix $\ME_{i,j}$ describes the relative pose transform between two cameras defined with a rotation matrix $\MR$ and translation $\vt$ up to scale as:
\begin{equation}
    \label{eq:essential_matrix_pose}
    \ME_{i,j} = \MR [\vt]_\times.
\end{equation}

Decomposing $\ME$ to recover $\vt$ and $\MR$ yields 
four solutions, of which only 
one satisfies the cheiral inequalities~\cite{Hartley98ijcv-chirality} which in summary state that triangulated points must lie in front of the two cameras.
Since eq.~\eqref{eq:essential_matrix_points} does not restrict scale, $\ME_{i,j}$ along with a point $\vr_{i}$ constrains 
a corresponding point $\vr_{j}$ in $I_j$ to a line known as the \emph{epipolar} line.

\myParagraph{Problem statement} 
Our objective is to determine if a given position estimate $\vxx_{est}$ is within some acceptable error bound, $\epsilon$, from the true position $\vxx_{gt}$:
\begin{equation}
    \label{eq:basic_condition}
    \| \vxx_{est} - \vxx_{gt} \| < \epsilon.
\end{equation}

All we assume are available is the position estimate $\vxx_{est}$, the sensor image $I_{gt}$, 
and a NeRF model whose weights are trained on a scene containing $I_{gt}$.  %

\section{\nameP}
\label{sec:contribution_pnp}

Here we present a simple yet effective method to estimate the correctness of a pose estimate using NeRF. We leverage 
NeRF to render a pair of stereo images to perform PnP. We first render an image $I_{est}$ at the estimated pose $\vxx_{est}$. 
Since the true pose $\vxx_{gt}$ is by definition the camera position corresponding to $I_{gt}$, the correctness constraint in eq.~\eqref{eq:basic_condition} 
can be satisfied by showing 
that the metric offset between $\vxx_{gt}$ and $\vxx_{est}$ is less than $\epsilon$. 
Towards this goal, we render a second image 
$I_{right}$ at $\vxx_{right}$ by translating $2\epsilon$ to the right with respect to $\vxx_{est}$. The image pair $I_{est}$ and $I_{right}$ whose poses 
are both known can then be used as a classical stereo pair of images. 
We compute the optical flow between these two images using RAFT~\cite{Teed20arxiv-RAFT} and use good features to track~\cite{shi94} 
to get sparse optical flow from RAFT's dense optical flow field. Likewise, we find the correspondences between $I_{est}$ and $I_{gt}$ for the same sparse points with RAFT. We then triangulate the 3D 
location of the sparse points by knowing $\vxx_{est}$ and $\vxx_{right}$ and finally apply PnP with RANSAC~\cite{Fischler81} to estimate the transform $\hat{\vxx}^{est}_{gt}$ between $\vxx_{est}$ and the unknown $\vxx_{gt}$. Our level 
of confidence in the accuracy of $\vxx_{est}$ is then estimated as follows:

\begin{equation}
    \label{eq:pnp_score}
    \probP( \| \hat{\vxx}^{est}_{gt} \| < \epsilon).
\end{equation}

We model $\| \hat{\vxx}^{est}_{gt} \|$ as a random variable whose mean value is the estimated position from PnP and standard deviation 
is manually selected. We will show in \cref{sec:experiments} the effectiveness of \nameP despite its simplicity.

\section{\nameL}
\label{sec:contribution_light}

\begin{figure*}
    \centering
    \begin{subfigure}[t]{0.228\textwidth}
        \centering
        \includegraphics[height=2.75in]{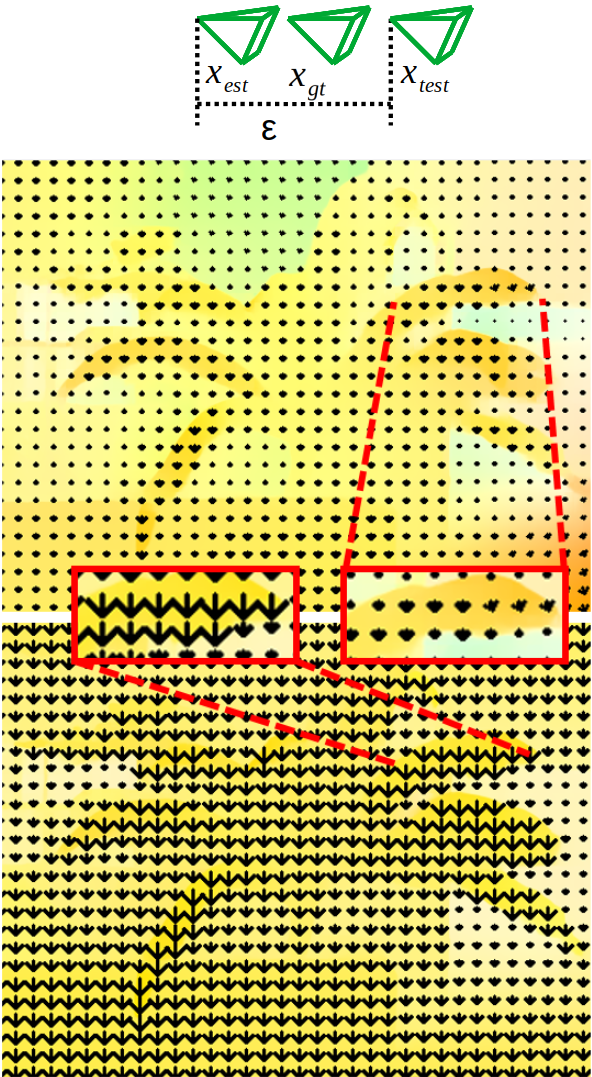}
        \caption{True position error is 1.9 cm. The flow field should allow for concluding that the pose estimate is correct with high confidence. 
        Order of camera positions shown above.}
        \label{fig:flow_cert}
    \end{subfigure}
    ~
    \begin{subfigure}[t]{0.228\textwidth}
        \centering
        \includegraphics[height=2.75in]{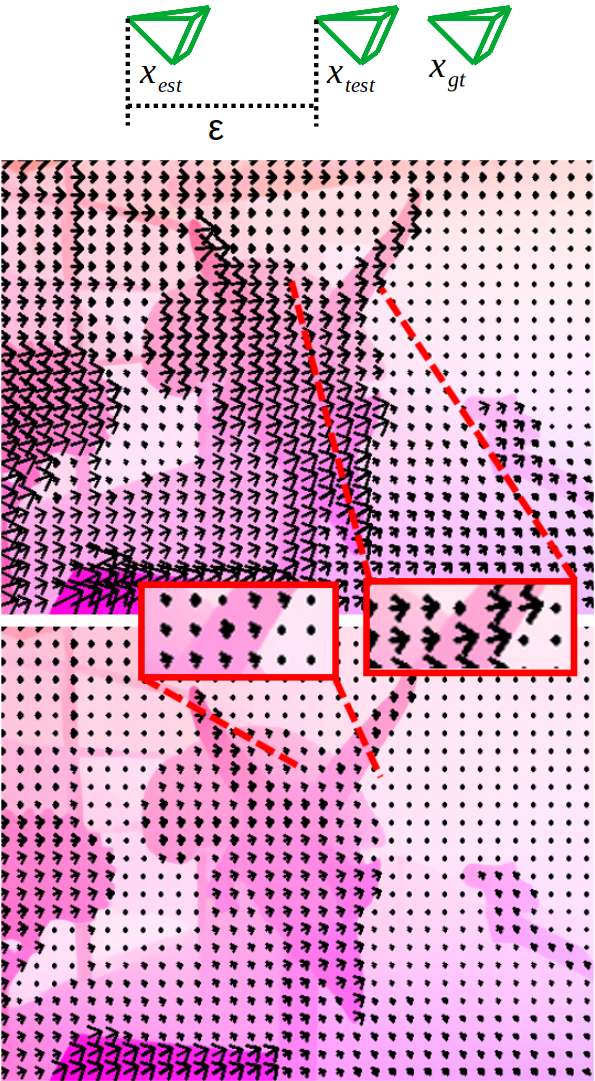}
        \caption{True position error is 10 cm. The flow field should allow for concluding that the pose estimate is incorrect with high confidence.
        Order of camera positions shown above.}
        \label{fig:flow_nocert}
    \end{subfigure}
    ~
    \begin{subfigure}[t]{0.228\textwidth}
        \centering
        \includegraphics[height=2.36in]{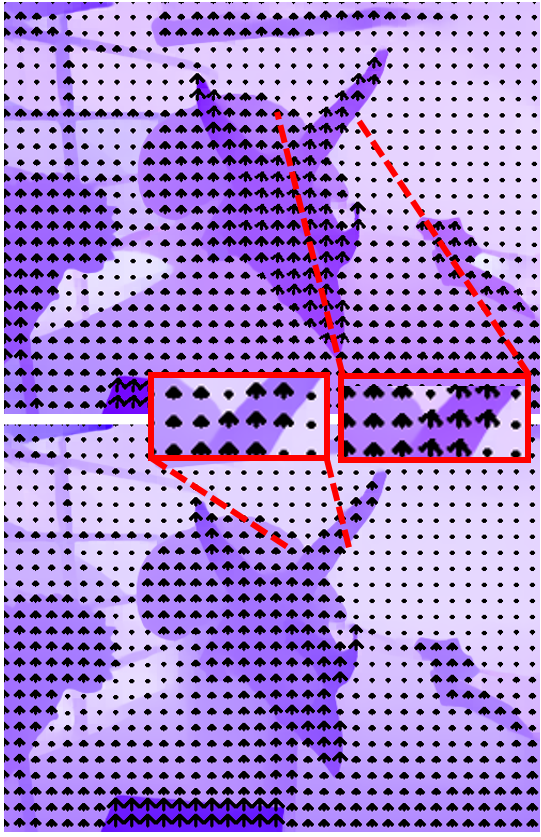}
        \caption{True position error is 5.6 cm. Pose can potentially be verified as correct, but should be done with low confidence.}
        \label{fig:flow_uncertain}
    \end{subfigure}
    ~
    \begin{subfigure}[t]{0.228\textwidth}
        \centering
        \includegraphics[height=2.36in]{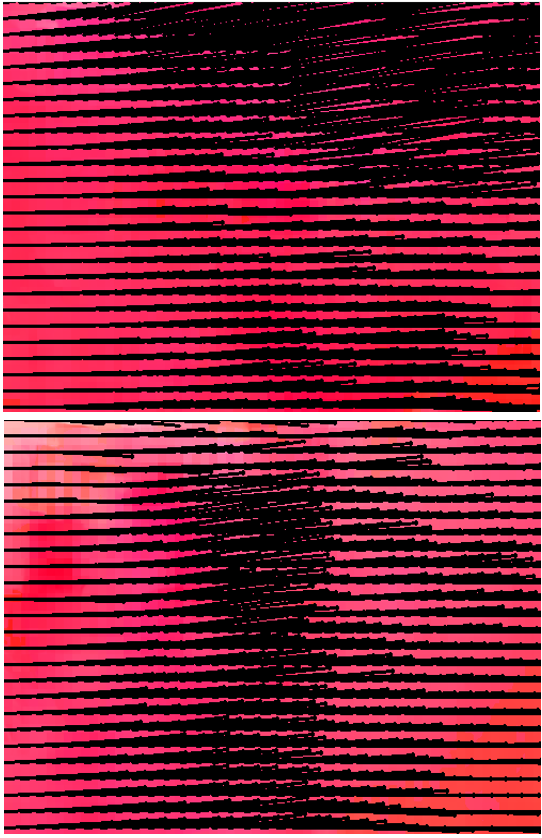}
        \caption{True position error is much larger than $\epsilon$ such that there are no clear correspondences between images.}
        \label{fig:flow_bad}
    \end{subfigure}
    \caption[Example conditions \nameL could encounter]{Example of optical flow between $I_{est}$, $I_{gt}$, and $I_{test}$ for pose estimates with a correctness condition of $\epsilon = 5 cm$. 
    \textbf{Top row}: optical flow between $I_{est}$ and $I_{gt}$. \textbf{Bottom row}: optical flow between $I_{est}$ and $I_{test}$.}
    \label{fig:overview}
\end{figure*}

\nameL can be divided into three phases (\cref{fig:diagram}): computing the relative offset between $\vxx_{est}$ and $\vxx_{gt}$ up to scale, 
selecting a test position $\vxx_{test}$ distance $\epsilon$ from $\vxx_{est}$ and co-linear with the latter two poses, 
and computing a quality of assurance that eq.~\eqref{eq:basic_condition} is met by using an application of the 
cheiral constraint. In particular, we leverage the fact that given three images from camera poses that are co-linear and with the same rotation, their order along the line they belong to can be determined by comparing the optical flow fields between them. 
For this arrangement, the flow fields between $I_{est}$ and $I_{gt}$ will be in the same direction as the flow field between $I_{est}$ and $I_{test}$, 
and the order of the three positions $\vxx_{est}$, $\vxx_{gt}$, and $\vxx_{test}$ can be estimated by comparing the magnitude of corresponding 
vectors between the two flow fields. If $\vxx_{gt}$ falls between $\vxx_{est}$ and $\vxx_{test}$ in such ordering, we can conclude that the error of $\vxx_{est}$ is less than $\epsilon$.

\myParagraph{Examples and intuition} \Cref{fig:overview} shows four example conditions that \nameL could potentially encounter. In \cref{fig:flow_cert} the flow field 
should provide confidence that the estimated pose is correct. First, the two optical flow fields have similar directions (and hence the same epipole) 
which validates our assumption of $\vxx_{est}$, $\vxx_{gt}$, and $\vxx_{test}$ being co-linear. Secondly, the magnitude of the optical flow 
between $I_{est}$ and $I_{test}$ (which have camera centers $\epsilon$ apart) is significantly greater than the corresponding flow between $I_{est}$ and $I_{gt}$ meaning that $\vxx_{gt}$ falls between $\vxx_{est}$ and $\vxx_{test}$ and hence the estimated pose is 
within $\epsilon$ of the true pose. In \Cref{fig:flow_nocert}, the estimate can 
safely be labeled as incorrect as there is clear evidence from the flow field that the flow 
between $I_{est}$ and $I_{test}$ is less in magnitude than the flow field between $I_{est}$ and $I_{gt}$ and again that the three perspectives are co-linear. 
\Cref{fig:flow_uncertain} on the other hand does not allow drawing strong conclusions. In this case there should be reduced confidence in the correctness assessment 
 as the flow field is roughly the same and differences may be only the result of noise. \Cref{fig:flow_bad} should be determined to be an incorrect pose but 
because of a different reason than \cref{fig:flow_nocert} --- here a cue that the pose is wrong is because no clear correspondences can be found between $I_{est}$ and $I_{gt}$. 

\subsection{\mbox{Computing Relative Error Direction of Position Estimate}}
\label{sec:rel_offset}

We use NeRF along with $\vxx_{est}$
to render an image $I_{est}$ which is the image that the camera would see if its center were at $\vxx_{est}$.
We use RAFT to compute the dense optical flow between $I_{gt}$ and $I_{est}$, 
and use good features to track~\cite{shi94} to
extract a set $\mathcal{R}$ of $n$ pixel coordinates $\vr_{{est}}$ 
to get sparse optical flow between the two images. 

We use the 5-point algorithm~\cite{Nister04pami} with RANSAC 
to determine the essential matrix $\ME_{est,gt}$.
RANSAC attempts to search for a $\hat{\ME}_{est,gt}$ such that a maximum number of points in $\mathcal{R}$ 
have sampson distance 
(a geometric constraint related to eq.~\eqref{eq:essential_matrix_points}) less than $\delta$. 
In short, the sampson distance~\cite{Sampson82cgip-sampsonDist} is an 
approximation of error to the epipolar line for two corresponding points. 
The unique solution to extracting the relative 
position $\hat{\bar{\vxx}}^{est}_{gt}$ 
up to scale from $\hat{\ME}_{est,gt}$ is found using the cheiral constraints with maximum consensus. 
Any points whose correspondence are not part of the maximum consensus or whose sampson distance is larger than $\delta$ 
are removed from the set of inliers $\mathcal{R}$ reducing the set of points to 
$\vr_{{est}} \in \mathcal{R'} \in \mathcal{R}$ where $n'$ is the number of points currently labeled as inliers.

\subsection{\mbox{Computing Location of Test Position}}
\label{sec:test_pose}

We now calculate a test position, $\vxx_{test}$, that is distance $\epsilon$ from $\vxx_{est}$ and co-linear with $I_{est}$ and $I_{gt}$:
\begin{equation}
    \label{eq:xtest}
    \vxx_{test} = \vxx_{est} + \epsilon \hat{\bar{\vxx}}^{est}_{gt}.
\end{equation}

The correctness condition in~\eqref{eq:basic_condition} can now be stated as:
\begin{equation}
    \label{eq:xtest_condition}
    \| \vxx_{est} - \vxx_{gt} \| < \| \vxx_{est} - \vxx_{test} \| = \epsilon
\end{equation}
where the exact pose of $\vxx_{est}$ and $\vxx_{test}$ are known and chosen to be $\epsilon$ apart. 
Note that since the positions are collinear by construction, the condition $\| \vxx_{est} - \vxx_{gt} \| < \| \vxx_{est} - \vxx_{test} \|$ is the same as requiring that these positions are ordered as $\vxx_{est},\vxx_{gt},\vxx_{test}$ along the line they belong to.
We render a new image $I_{test}$ at $\vxx_{test}$ using NeRF.

\subsection{\mbox{Determining the Confidence Score}}
\label{sec:confidence_of_validation}

We again use RAFT to compute the dense optical flow, this time between $I_{est}$ and $I_{test}$ and 
get sparse optical flow $\hat{\vv}(\vr_{est})_{I_{est},I_{test}}$ for coordinates $\vr_{{est}} \in \mathcal{R'}$. 

We now consider several properties given our particular choice of $\vxx_{test}$. 
The first is that it is unnecessary to compute $\ME_{est, test}$ as we directly know it without error from the true poses of $\vxx_{est}$ and $\vxx_{test}$. 
Furthermore, it is simply the same as our estimate of $\ME_{est, gt}$ since $\vxx_{est}$, $\vxx_{gt}$, and $\vxx_{test}$ are 
aligned and co-linear. This is summarized in the following relation:

\begin{equation}
    \label{eq:three_essential_matricies}
    \hat{\ME}_{est,gt} = \ME_{est, test}.
\end{equation}

Determining whether eq.~\eqref{eq:xtest_condition} is satisfied now reduces to solving an image ordering problem for $I_{est}, I_{gt}, I_{test}$ outlined 
visually in \cref{fig:diagram}. 
If $\mathcal{R'}$ contains only true, noiseless inliers, the image ordering problem could now be solved using an application of the cheiral constraint:
\begin{equation}
    \label{eq:noiseless_good_pose}
    \begin{split}
    \| \vxx_{est} - \vxx_{gt} \| < \epsilon \iff \\
     \forall \vr_{est} \in \mathcal{R'}, \| \vv(\vr_{est})_{I_{est}, I_{gt}} \| < \| \vv(\vr_{est})_{I_{est}, I_{test}} \|
    \end{split}
\end{equation}

Equation~\eqref{eq:noiseless_good_pose} %
states that for noiseless optical flow fields, the condition 
of correctness in 
\eqref{eq:xtest_condition} implies the optical flow vector 
relating a point $\vr_{{est}}$ to its corresponding point in $I_{gt}$ should be 
of less magnitude than the flow vector relating $\vr_{{est}}$ to its corresponding point in $I_{test}$. 
The two corresponding vectors are in the same direction since the three poses are co-linear and hence the points $\vr_{{gt}}$ and $\vr_{{test}}$ 
corresponding to $\vr_{{est}}$ are bound to the same epipolar line.

However, in the presence of noise and false inliers, we must consider the possibility that 
the epipolar constraint in eq.~\eqref{eq:essential_matrix_points} is not exactly satisfied and hence $\mathcal{R'}$ may contain 
false inliers, the location of points $\vr_{gt}$ and $\vr_{test}$ along the
epipolar line $\hat{\ME}_{est,gt} \tilde{\vr}_{I_{est}}$ are perturbed by noise, and that $\hat{\ME}_{est,gt}$ differs from $\ME_{est,gt}$.
A primary source of error in our proposed monitoring method is the calculation of optical flow. 
Our estimate of the optical flow for any particular point can be expressed as follows:

\begin{equation}
    \label{eq:flow_error}
    \hat{\vv}(\vr_{i})_{ij} = \vv(\vr_{i})_{ij} + \vo_{ij} + \vgamma_{ij}
\end{equation}
where $\| \vgamma_{ij} \| \leq \delta$ and $\vo_{ij}$ is 0 if $\hat{\vv}(\vr_{i})_{ij}$ is an inlier with sampson distance less than $\delta$.
Otherwise, in the case of an outlier, $\vo_{ij}$ is any arbitrary value such that $\hat{\vv}(\vr_{i})_{ij}$ can exist at any location in the image. 
By computing the sampson distance of each $\hat{\vv}(\vr_{est})_{I_{est},I_{test}}$ w.r.t. $\hat{\ME}_{est,gt}$, we can filter out points with error larger 
than $\delta$. Note this does not check for error along the epipolar line. We additionally filter out points which are not part of the cheiral set of maximum consensus. 
We again prune out any points whose correspondences have been labeled as outliers from a set of size $n'$ to a set of $n''$, i.e., 
$\vr_{{est}}  \in \mathcal{R''} \in \mathcal{R'}$. 
Lastly, we project all of $\hat{\vr}_{I_{gt}}$ and $\hat{\vr}_{{test}}$ to the epipolar line defined by $\hat{\ME}_{est,gt} \tilde{\vr}_{I_{est}}$
yielding $\hat{\dot{\vr}}_{{gt}}$ and $\hat{\dot{\vr}}_{{test}}$ such that pairs of corresponding points satisfy eq.~\eqref{eq:essential_matrix_points}.

\myParagraph{Computing the confidence score}
Now we must estimate the confidence, $q$, that the optical flow for corresponding points between $I_{est}$ to $I_{test}$ is 
greater than the ones between $I_{est}$ to $I_{gt}$, i.e.,
$\| \vv(\vr_{est})_{I_{est}, I_{gt}} \| < \| \vv(\vr_{est})_{I_{est}, I_{test}} \|$. Using the optical flow vectors from 
$\vr_{est}$ to the 
projected points $\hat{\dot{\vr}}_{{gt}}$ and $\hat{\dot{\vr}}_{{test}}$ we define the following confidence score:
\begin{equation}
    \label{eq:verification_score}
    q = \frac{1}{n''} \sum_{i=1}^{n''} \probP(\| \hat{\dot{\vv}}(\vr_{est})_{I_{est}, I_{gt}} \| < \| \hat{\dot{\vv}}(\vr_{est})_{I_{est}, I_{test}} \|).
\end{equation}

Explicitly, the confidence score in \eqref{eq:verification_score} is computed using the Normal CDF with a user-specified variance $V$. Standard deviation is set to a reasonable value of pixel error (e.g., 0.5). 
As a results, we rewrite \eqref{eq:verification_score} as:
\begin{equation}
    \label{eq:cdf}
    q = \frac{1}{n''} \sum_{i=1}^{n''} 
    \Phi\left(\frac{\hat{\dot{\vv}}(\vr_{est})_{I_{est}, I_{test}} - \hat{\dot{\vv}}(\vr_{est})(\vr_{est})_{I_{est}, I_{gt}}}{\sqrt{V[\hat{\dot{\vv}}(\vr_{est})_{I_{gt}}]}}\right)
\end{equation}
where $\Phi$ is the Normal CDF.
The confidence score mimics a probability, however due to simplifying assumptions such 
as approximating optical flow uncertainty and potential errors in computing the essential matrix, we do not claim it to be a true probability. %

\section{Experiments}
\label{sec:experiments}

We now present results of running \name-PnP and \name-Light on three types of environments ranging from small-scale indoor scenes 
to a rocket trajectory spanning 8 km. 
For all experiments, we use torch-ngp~\cite{Tang22-torch-ngp} as our NeRF model. 
To get experimental sensor images we use randomly selected images from the NeRF training set. 
For each image, we generate a pose estimate to be 
checked for correctness by adding a random offset to the corresponding ground-truth position. 
To get a diverse distribution of correct and incorrect poses, we randomly selected either a low or high error regime when generating offsets.

In addition to comparing the two proposed methods, we include a simple baseline method that we refer to as Disparity Check.
For this, we simply compute the optical flow between $I_{est}$ and $I_{gt}$ and determine the mean disparity from 
sparse flow. 
A naive approach is to assume low disparity means a correct pose estimation whereas a high disparity points to an 
incorrect pose. We use a folded normal distribution which computes a confidence level of correctness given a mean disparity. 
All experiments use a standard deviation of 4 pixels for the folded normal distribution.
Since this method makes no efforts to handle scale ambiguity, we will show that it does not generalize well across varying scene size.

We pick a 0.5 cutoff confidence level for each method to
estimate if the pose is correct or not. To show the generalizability of \name, for all experiments we use the same standard deviation in \eqref{eq:cdf} for \nameL (0.5 pixels)
and the same standard deviation for \nameP in \eqref{eq:pnp_score}. 
Likewise, the same RANSAC, RAFT, and good features to track parameters are used for all experiments.

\subsection{\mbox{LLFF dataset}}
\label{sec:experiments_llff}

\myParagraph{Setup}
We first evaluate \name on 4 scenes (Fern, Fortress, Horns, and Room) from the LLFF
dataset~\cite{Mildenhall19tog-llff}. We pick 250 randomly selected views from the training set of images for each scene to 
serve as the sensor image $I_{gt}$ and 
for each image randomly generate a choice for $\vxx_{est}$. 
We downscale $I_{gt}$ to 504$\times$378 and render the same resolution images when using NeRF.
For these 1000 tests, we set $\epsilon$ to be 5 cm.

\begin{figure}[H]
    \includegraphics[width=0.49\textwidth, height=\plotheight]{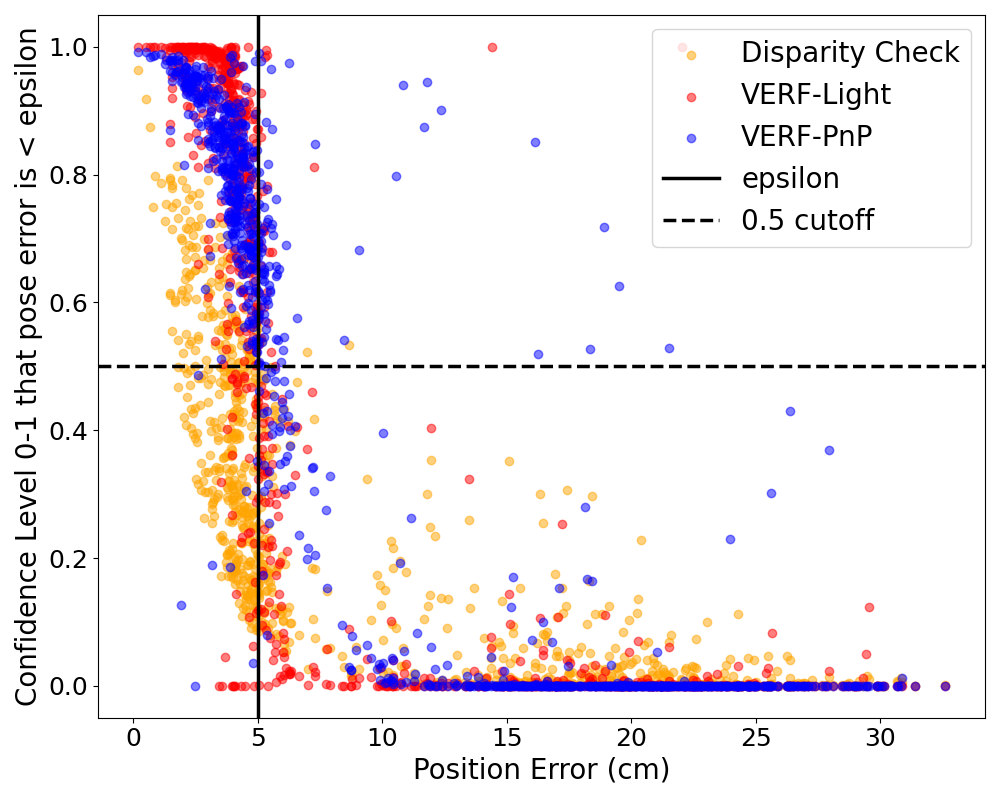}
    \caption{
    \name confidence level that for 1000 randomly sampled position estimates for LLFF scenes error is less than $\epsilon=5$ cm
    \label{fig:llff_exp} \vspace{-3mm}}
\end{figure}

\myParagraph{Results}
In \cref{fig:llff_exp} we show the level of confidence \name computed that the position error is less than $\epsilon$ compared to the actual position error for each test. 
As expected, confidence levels 
approach 1 as the position error is well less than $\epsilon$ and approach 0 when the position error is much greater than $\epsilon$. 
On a 3090 GPU, total time to produce a confidence score from \nameL is on average 0.4 seconds with 0.25 seconds of that used for \nerf rendering, and 
is on average 0.35 seconds for \nameP with the same time used for rendering since each method renders two \nerf images. 

A summary of results is provided in \cref{table:llff}. Similar performance is observed by \nameL and \nameP with most misclassifications 
occurring for pose estimates with errors near epsilon. 
Both methods outperform the Disparity Check baseline by a vast margin.

\begin{table}[H]
    \centering
    \scriptsize
    \begin{tabular}{ | l | l | l | l |}
        \hline
         & Disparity Check & \nameP & \nameL \\ \hline
        True Positives (423)& 146 & 415 & 381 \\ \hline
        True Negatives (577)& 572 & 494 & 545 \\ \hline
        False Positives& 5 & 83 & 32\\ \hline
        False Negatives& 277 & 8 & 42 \\ \hline
        \textbf{Total Correct}& \textbf{72\%} & \textbf{91\%} & \textbf{93\%} \\ \hline
    \end{tabular}
    \caption{Summary of results for all proposed methods on 1000 tests on LLFF dataset. Classification is made 
    with a 0.5 confidence score cutoff.}
    \label{table:llff}
\end{table}

\subsection{\mbox{A1 Quadruped}}
\label{sec:experiments_a1}

\begin{figure}[hbt]
    \includegraphics[width=0.47\textwidth]{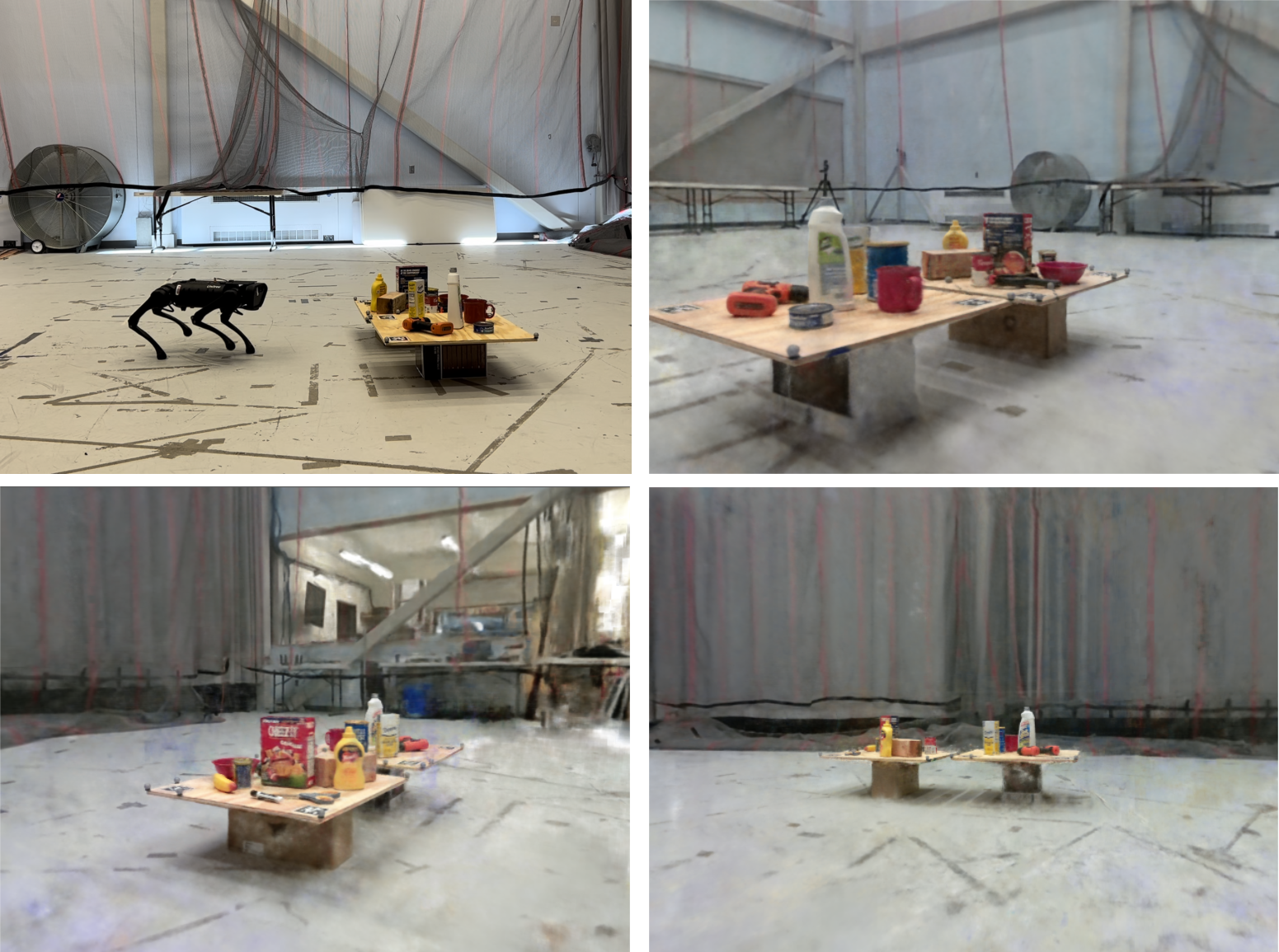}
    \caption[Generating a NeRF with an A1 quadruped robot]{
    A1 quadruped robot collecting monocular RGB data for NeRF training and \name evaluation (top left).
    Three example NeRF-rendered views using weights trained by camera data collected onboard an A1 robot.
    \label{fig:a1_nerf_robot} \vspace{-3mm}}
\end{figure}

\myParagraph{Setup}
We train a NeRF (\cref{fig:a1_nerf_robot}) using RGB images collected with a Realsense D455 camera mounted on a Unitree A1 quadruped robot (\cref{fig:a1_nerf_robot}). 
The robot transverses around a table at varying distances to the table in a motion capture room. 
Training images and sensor images are downscaled to $640 \times 360$. 
Ground-truth poses are estimated with \colmap \cite{schoenberger16cvpr-colmap}. 
To correct from the ambiguous scale 
from \colmap, we use vicon odometry to add metric scale to the poses. We again randomly select 1000 images with replacement 
from the dataset as sensor images and generate a random pose estimate for each image to be verified.

\begin{figure}[hbt]
\centering
    \includegraphics[width=0.49\textwidth, height=\plotheight]{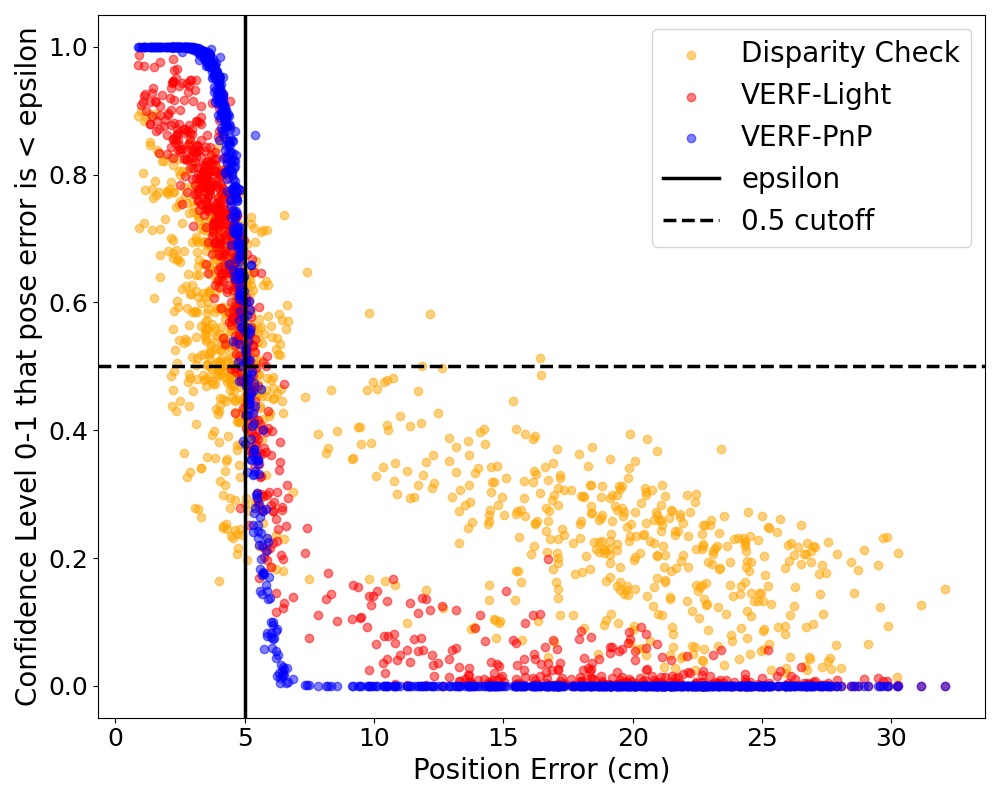}
    \caption[\name results for 1000 tests on A1 room scene]{
        \name confidence level that for 1000 randomly sampled position estimates of the A1's pose, error is less than $\epsilon=5$ cm
        \label{fig:a1_exp} \vspace{-3mm}}
\end{figure}

\myParagraph{Results}
We pick epsilon to be 5 cm and observe similar results as with the LLFF experiment 
with nearly all \name mistakes occurring for position errors near the value of epsilon. Results are summarized in \cref{table:a1} and shown visually in \cref{fig:a1_exp}. 
The Disparity Check baseline is shown to generalize poorly for different scale scenes as most of its errors are false negatives for the LLFF experiment 
whereas most of its errors are false positives for the A1 experiment. 

\begin{table}[H]
    \centering
    \scriptsize
    \begin{tabular}{ | l | l | l | l |}
        \hline
         & Disparity Check & \nameP & \nameL \\ \hline
         True Positives (421)& 304 & 418 & 411 \\ \hline
         True Negatives (579)& 513 & 561 & 551 \\ \hline
         False Positives& 66 & 18 & 28 \\ \hline
         False Negatives& 117 & 3 & 10 \\ \hline
        \textbf{Total Correct}& \textbf{82\%} & \textbf{98\%} & \textbf{96\%} \\ \hline
    \end{tabular}
    \caption[Summary of results for \nameL on 1000 tests on A1 dataset]{Summary of results for all proposed methods on 1000 tests of A1 robot dataset. Classification is made 
    with a 0.5 confidence score cutoff.}
    \label{table:a1}
\end{table}

\subsection{\mbox{Sub-Orbital Rocket}}
\label{sec:experiments_rocket}

\begin{figure}[hbt]
    \includegraphics[width=0.47\textwidth]{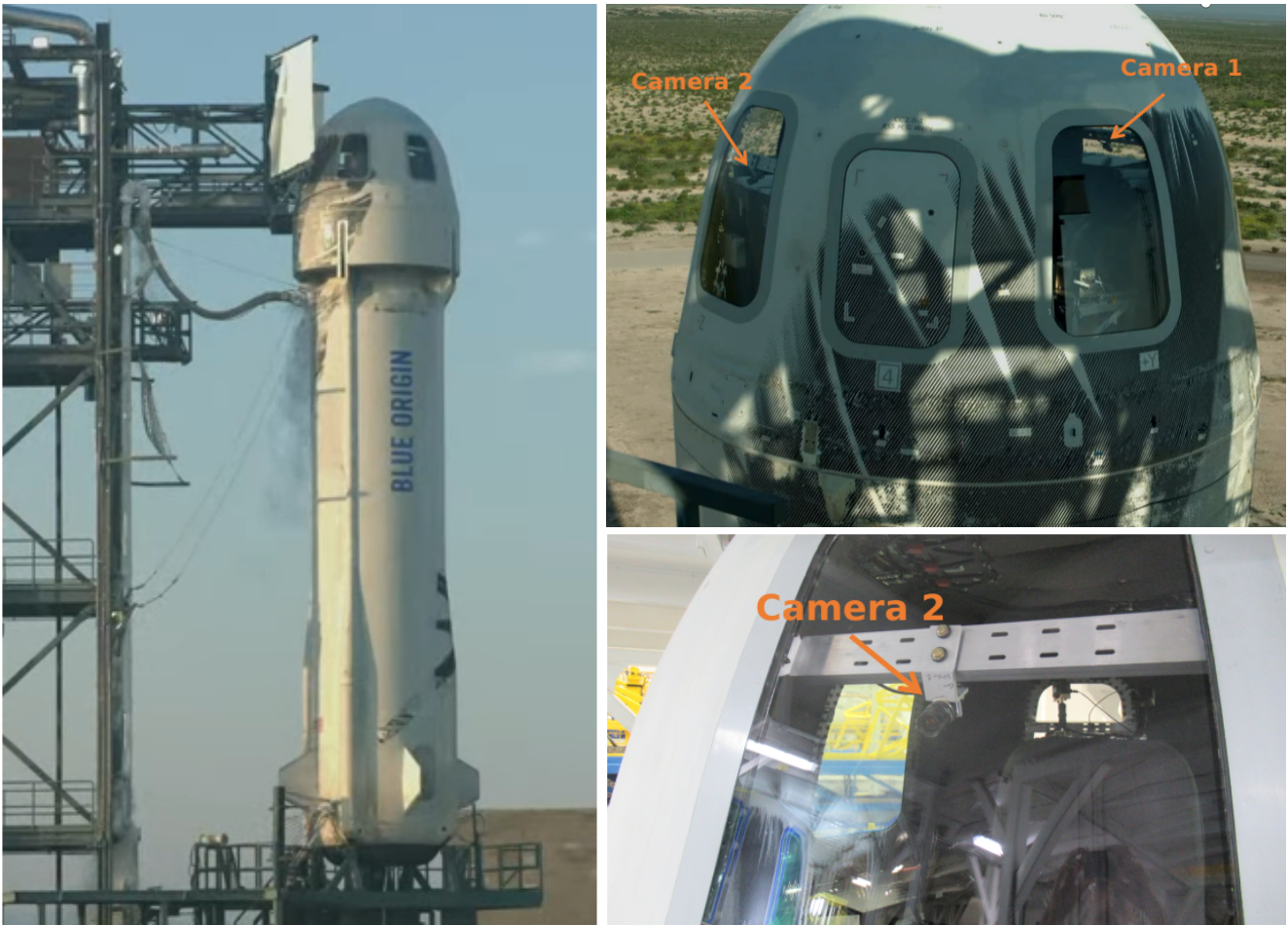}
    \caption{
    Sub-orbital rocket on launch pad used for data collection (left). Close view of camera mounted 
    inside the capsule window (bottom right) and a view of both cameras inside the capsule before 
    launch (top right). Images courtesy of Blue Origin.
    \label{fig:rocket} \vspace{-3mm}}
\end{figure}

\myParagraph{Setup}
Here we demonstrate the potential for \name to be used in a highly complex scenario such as for precision spacecraft 
navigation. This experiment uses data we collected for \cite{Maggio23scitech-IBAL} in which we mounted two cameras inside the capsule 
of Blue Origin's New Shepard rocket which point out the capsule windows towards the terrain (\cref{fig:rocket}).

We train on 140 images collected during the rocket's ascent from an altitude 
range of approximately 0.2 to 8 km above ground level  
during which the rocket reaches a speed up to 880 km/hr. We do not include data at higher altitudes as there was a 
mishap during flight NS-23 which triggered the capsule escape system.  The curious reader can refer to 
\cite{Maggio23scitech-IBAL} for more details of our flight data collection.

For simplicity, we train on images collected during the flight and use estimated poses from \colmap 
as ground truth. In practice, a NeRF could be trained from prior satellite maps as was done 
in~\cite{xiangli22eccv-bungeenerf}. Again, similar to the A1 experiment, \name is run on a scaled \nerf model 
and we provide metric scale to the \colmap reconstruction from 
ground truth poses of the training images --- in this case from GPS inside the rocket's capsule.

\begin{figure}[hbt]
    \includegraphics[width=0.47\textwidth]{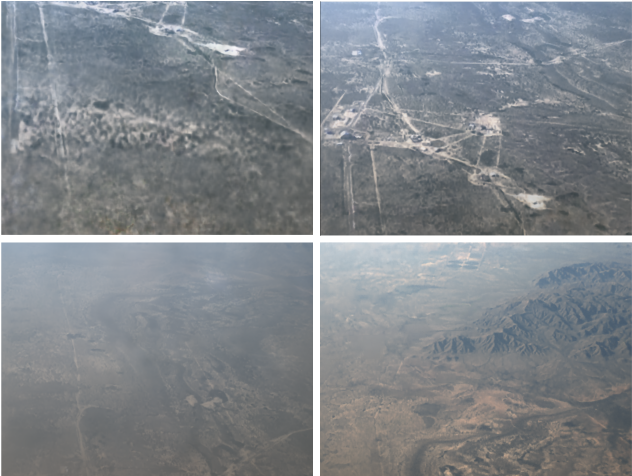}
    \caption{
    Example of four NeRF rendered views from sub-orbital rocket ascent from an altitude range of approximately 1 km to 8 km.
    \label{fig:rocket_nerf} \vspace{-3mm}}
\end{figure}

\begin{figure}[hbt]
    \includegraphics[width=0.49\textwidth]{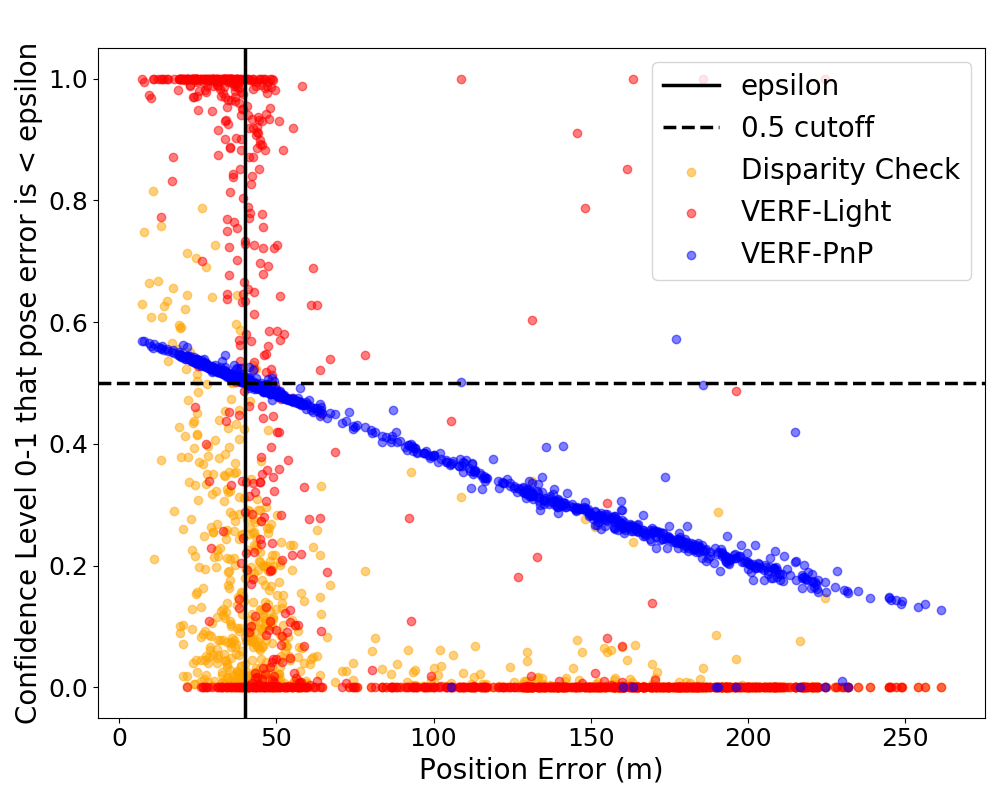}
    \caption{
    \name confidence level that for 1000 randomly sampled position estimates of the rocket's pose, error is less than $\epsilon=40$ m
    \label{fig:blue_exp} \vspace{-3mm}}
\end{figure}

\myParagraph{Results}
We pick 40 m for epsilon since this is on the order of typical spacecraft landing accuracy 
for planetary exploration~\cite{Johnson17-lvs}. A summary of results 
is shown in \cref{table:rocket} and visually in \cref{fig:blue_exp}. 
\nameP performs notably better than \nameL on this dataset which we believe to be caused by inaccuracies in the essential matrix 
estimation due to the scene being approximately planar at high altitudes. \Cref{sec:appendix} provides a study on the effects of error in 
the essential matrix on \nameL. 

\begin{table}[hbt]
    \scriptsize
    \centering
    \begin{tabular}{ | l | l | l | l | l |}
        \hline
         & Disparity Check & \nameP & \nameL\\ \hline
        True Positives (260)& 38 & 259 & 214 \\ \hline
        True Negatives (740)& 738 & 710 & 640 \\ \hline
        False Positives& 2 & 30 & 100 \\ \hline
        False Negatives& 222 & 1 & 46 \\ \hline
        Total Correct& \textbf{78\%} & \textbf{97\%} & \textbf{85\%} \\ \hline
    \end{tabular}
    \caption{Summary of results for all proposed methods on 1000 tests of rocket dataset. Classification is made 
    with a 0.5 confidence score cutoff.}
    \label{table:rocket}
\end{table}

\begin{figure}[H]
    \includegraphics[width=0.49\textwidth, height=5.9cm]{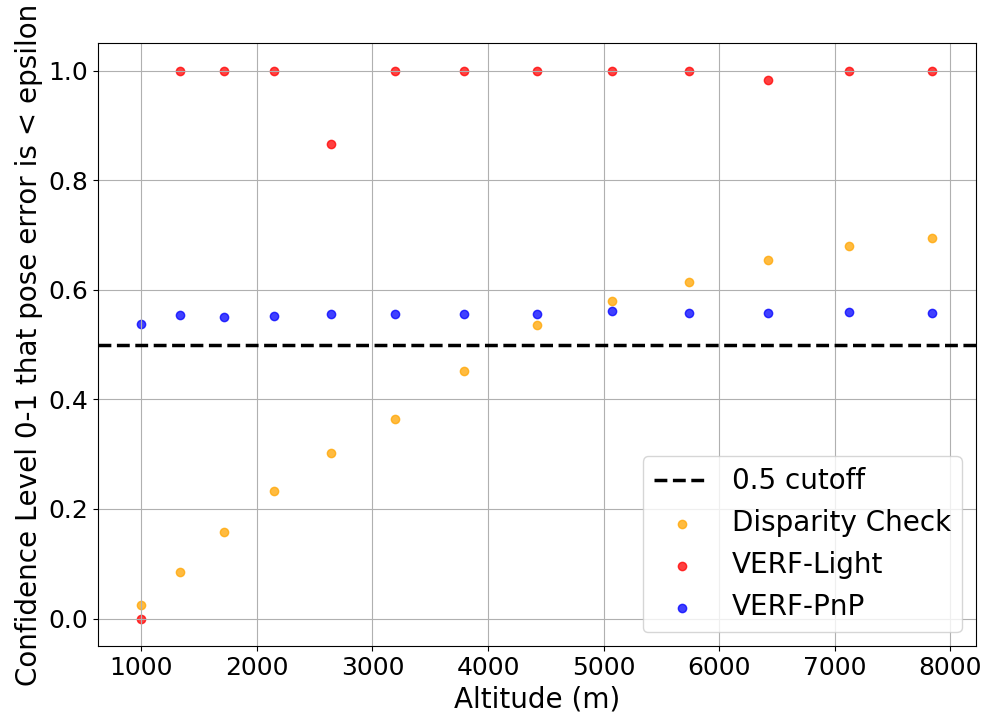}
    \caption{
    \name confidence level vs altitude for rocket dataset with fixed 15 m of position error. Epsilon is selected as 40 m. 
    Disparity Check is shown to not generalize with varying scene scale while \nameP and \nameL performance are independent 
    of the rocket's altitude.
    \label{fig:rocket_alt} \vspace{-3mm}}
\end{figure}

Additionally, as \name must perform well across a wide range of altitudes for the rocket dataset, we show that \nameP and \nameL 
perform well across all altitudes while Disparity Check does not generalize well. To further demonstrate this point, in \cref{fig:rocket_alt} we pick estimated 
poses with error 15 m (with epsilon again set to 40 m) and run all three methods on sequential images during launch. \Cref{fig:rocket_alt} 
shows that the performance of the Disparity Check is dependent on altitude (switching its decision from incorrect to correct after 4 km) 
while \nameP and \nameL perform consistently 
throughout the rocket's accent. 

\section{Conclusion}
\label{sec:conclusion}

We have presented two approaches (\nameP and \nameL) to leverage Neural Radiance Fields to monitor the correctness of a pose estimate acquired from a monocular camera. 
Our methods functions independently of how the pose is estimated (i.e., NeRF does not have to be used for pose estimation) 
and can provide a level of assurance in under half a second. 
Experiments have shown the effectiveness 
of \name on scene scales ranging from small rooms to kilometer-scale outdoor scenes. 
As a limitation, we remark that
\name is intended to be a local pose monitoring approach in the sense that if an arbitrarily large epsilon were used, it is possible 
for NeRF rendered images to be outside the range of the trained NeRF or fail to match features to the sensor image leading to a false 
assumption of an incorrect pose.  
\section*{Acknowledgement}
The authors gratefully acknowledge Jingnan Shi, Brett Streetman, and Ted Steiner for their help collecting data for our experiments.

\bibliographystyle{IEEEtran}

\clearpage

\section{Appendices}
\label{sec:appendix}

\subsection{Effects of Essential Matrix Error}
\label{sec:true_e}
\begin{figure}[hbt]
    \includegraphics[width=0.49\textwidth,height=\plotheight]{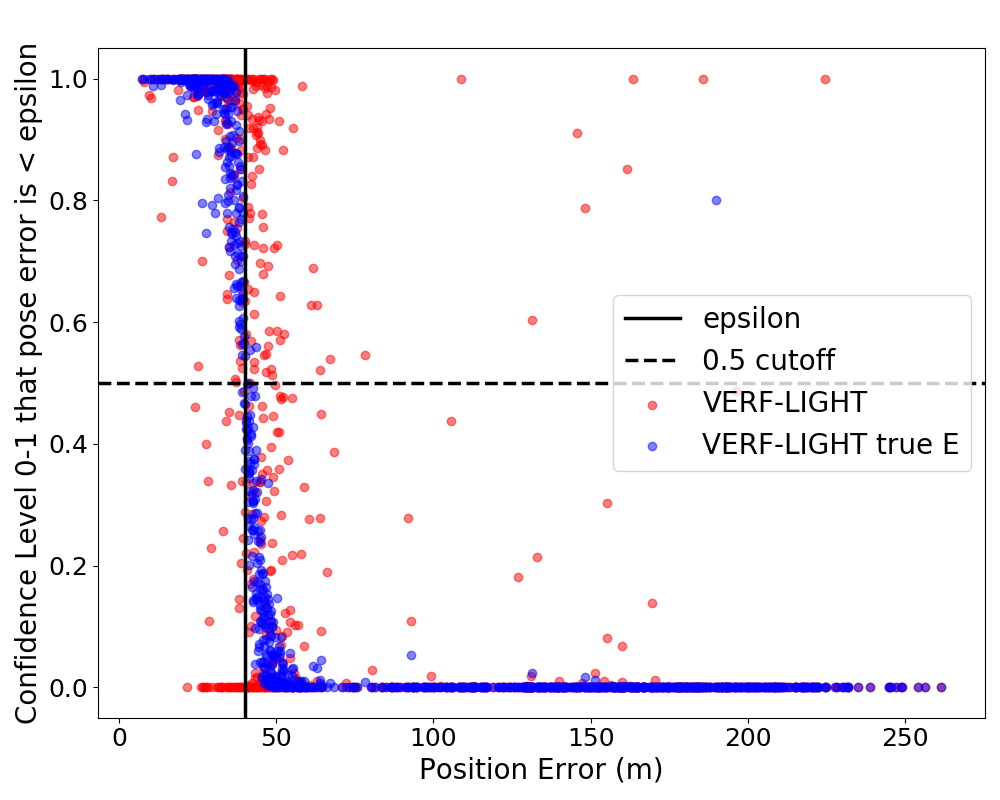}
    \caption{
    \name confidence level that for 1000 randomly sampled position estimates for rocket data that their error is less than $\epsilon=40 m$. 
    \nameL shown with and without the true essential matrix. 
    \label{fig:llff_true_e} \vspace{-3mm}}
\end{figure}

Here we study the effects of the accuracy of the essential matrix estimation for \nameL. We repeat each of the three experiments 
with the same setup except we now provide \nameL with the true essential matrix $\ME_{est, gt}$. \Cref{table:results_true_e} shows 
notable improvements on all experiments with \nameL correctly classifying 99\% of pose estimates. The most significant improvement 
is on the rocket dataset. Not only does accuracy go from 86\% to 99\% but as shown in \cref{fig:llff_true_e}, the confidence levels 
follow a cleaner distribution. We believe this to be caused by the approximate planar scene from high altitudes. This study thus shows the 
potential to improve \nameL with a more effective essential matrix estimation method. 

\begin{table}[hbt]
    \centering
    \scriptsize
    \begin{tabular}{ | l | l | l | l |}
        \hline
         & LLFF & A1 & Rocket \\ \hline
        True Positives& 415 & 420 & 255 \\ \hline
        True Negatives& 574 & 567 & 736 \\ \hline
        False Positives& 3 & 12 & 4 \\ \hline
        False Negatives& 8 & 1 & 5 \\ \hline
        \textbf{Total Correct}& \textbf{99\%} & \textbf{99\%} & \textbf{99\%}\\ \hline
    \end{tabular}
    \caption{Summary of results on running \nameL on all experiments using the true essential matrix.}
    \label{table:results_true_e}
\end{table}

\subsection{Estimating Metric Error with PnP}
\label{sec:pose_pnp}

\begin{figure}[H]
    \includegraphics[width=0.49\textwidth]{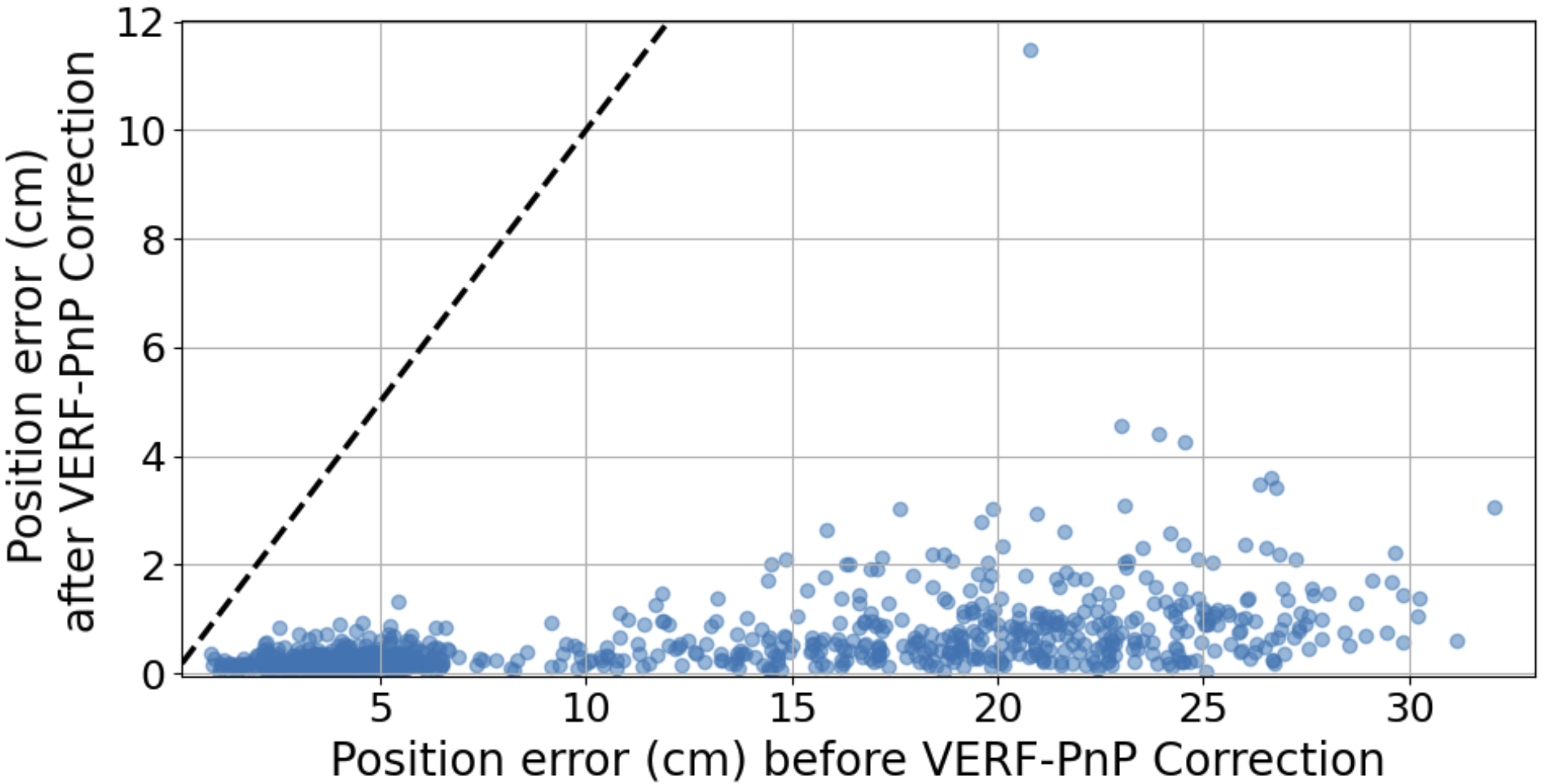}
    \caption{
        Position errors before and after being corrected using the position error estimate from \nameP. Results shown for 
        1000 tests on the A1 dataset.
    \label{fig:a1_corrected_pnp} \vspace{-3mm}}
\end{figure}

\begin{figure}[H]
    \includegraphics[width=0.49\textwidth]{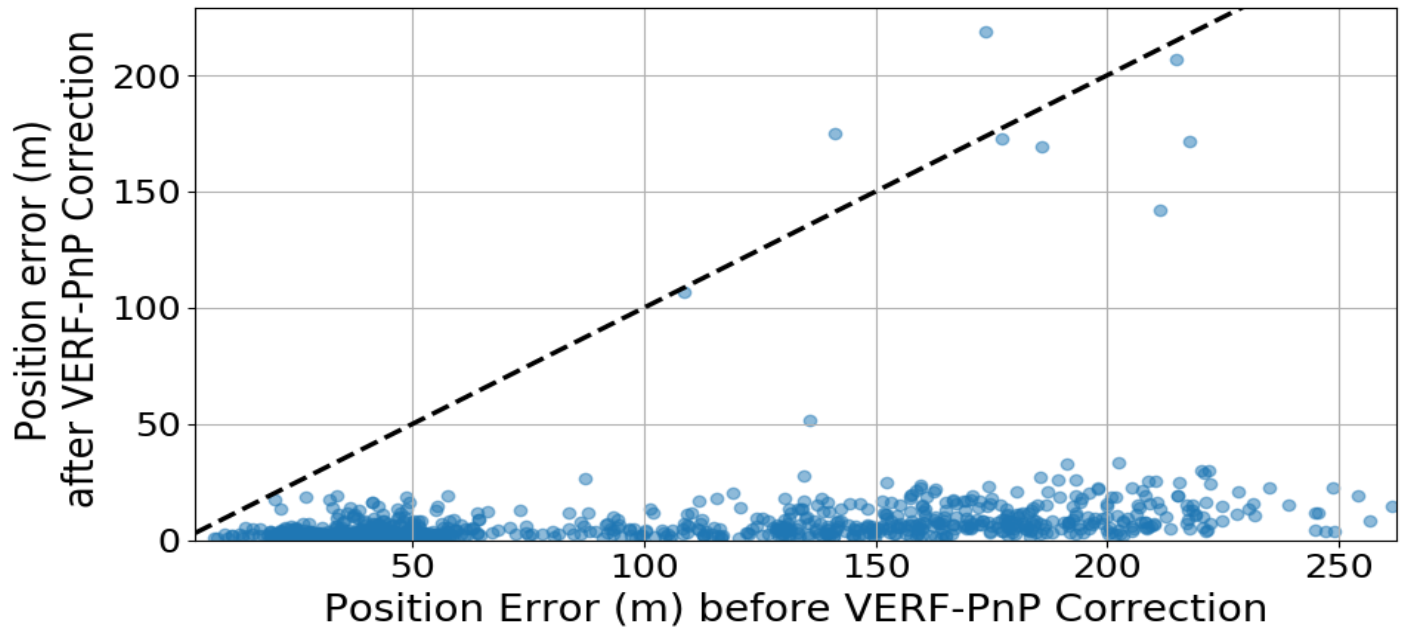}
    \caption{
        Position errors before and after being corrected using the position error estimate from \nameP. Results shown for 
        1000 tests on the Rocket dataset.
    \label{fig:rocket_corrected_pnp} \vspace{-3mm}}
\end{figure}

A logical question to pose is since PnP can estimate metric distance, how well can \nameP estimate the true error instead of just 
estimating correctness with respect to an epsilon threshold. With an estimate of the true error, the pose estimate can then be corrected. 
\Cref{fig:a1_corrected_pnp,fig:rocket_corrected_pnp} show position errors before an after being corrected in this fashion 
by \nameP with errors decreasing by an order of magnitude.
For each experiment there were a small number of pose estimates omitted 
(1 for \cref{fig:a1_corrected_pnp} and 9 for \cref{fig:rocket_corrected_pnp}) as PnP diverged. 
One potential option to automatically check and prevent this is to only accept the updated pose if 
\nameL predicts that the corrected pose is less than epsilon.

 \end{document}